\documentclass[journal]{IEEEtai}

\usepackage{cite}
\usepackage{amsmath,amssymb,amsfonts,amsthm}
\usepackage{algorithm}
\usepackage{algorithmic}
\usepackage{textcomp}
\usepackage{graphicx}
\usepackage{textcomp}
\usepackage{xcolor}
\usepackage{cite}
\usepackage{cases}
\usepackage{mathrsfs}
\usepackage{amscd}
\usepackage{graphicx}  
\usepackage{epstopdf}
\usepackage{subfigure}
\usepackage{diagbox}
\usepackage{multirow}
\usepackage{booktabs}
\usepackage{hyperref}
\renewcommand \thetable{\arabic{table}}
\usepackage{epstopdf}
\usepackage{bm}
\usepackage{booktabs}
\usepackage{multirow}
\usepackage{float}
\usepackage{stfloats}
\allowdisplaybreaks

\newcommand{\bfa}{\mathbf{a}}
\newcommand{\bfs}{\mathbf{s}}
\newcommand{\bfr}{\mathbf{r}}
\newcommand{\bfo}{\mathbf{o}}
\newcommand{\calP}{\mathcal{P}}
\newcommand{\calO}{\mathcal{O}}
\newcommand{\calQ}{\mathcal{Q}}
\newcommand{\calS}{\mathcal{S}}
\newcommand{\calA}{\mathcal{A}}
\newcommand{\calR}{\mathcal{R}}

\begin{document}


\title{A Convolution and Attention Based Encoder for Reinforcement Learning under Partial Observability}

\author{Wuhao Wang and Zhiyong Chen*
\thanks{W. Wang and Z. Chen are with the School of Engineering, The University of Newcastle, Callaghan, NSW 2308, Australia.}
\thanks{Corresponding author: Zhiyong Chen (e-mail: zhiyong.chen@newcastle.edu.au).}
}



\maketitle

\begin{abstract}
Partially Observable Markov Decision Processes (POMDPs) remain a core challenge in reinforcement learning due to incomplete state information. We address this by reformulating POMDPs as fully observable processes with fixed-length observation histories as augmented states. To efficiently encode these histories, we propose a lightweight temporal encoder based on depthwise separable convolution and self-attention, avoiding the overhead of recurrent and Transformer-based models. Integrated into an actor–critic framework, our method achieves superior performance on continuous control benchmarks under partial observability. More broadly, this work shows that lightweight temporal encoding can improve the scalability of AI systems under uncertainty. It advances the development of agents capable of reasoning robustly in real-world environments where information is incomplete or delayed.
\end{abstract}

\begin{IEEEImpStatement}
This work introduces a lightweight reinforcement learning approach for decision-making under partial observability. By combining depthwise convolution and attention mechanisms, our method surpasses recurrent networks while maintaining a comparable model size, thus making advanced AI control more accessible and reliable in sensor-limited scenarios. The approach not only demonstrates practical efficiency but also offers a clearer pathway toward scalable sequence modeling in reinforcement learning. In addition, the study provides new theoretical insights by showing how the parallel encoding of attention can, under mild assumptions, simplify partially observable problems into more tractable Markov decision process, thereby broadening the foundation for future research.
\end{IEEEImpStatement}



\section{Introduction}
\IEEEPARstart{R}einforcement Learning (RL) has been widely applied in diverse domains, including game strategy optimization\cite{Wurman2022outracing, Xu2022perceiving}, scientific research \cite{Payal2024Deep, Li2024Deep}, and complex system control \cite{degrave2022magnetic,chen2023reinforcement}. Its success demonstrates great potential for policy optimization. However, RL is not without limitations. Most existing algorithms rely on the agent having full access to the system state, whereas real-world applications are often partially observable, providing only partial measurements due to sensor and modeling limitations. In such cases, a single observation is insufficient to recover the latent state, but sequences of past observations can contain the missing information. In such cases, it is well established that a single observation is insufficient to recover the latent state, and that leveraging history information is a standard way to mitigate partial observability \cite{Bakker2001Reinforce, Hausknecht2015drqn}. 

\begin{figure}[th]
\centering
\includegraphics[width=0.9\linewidth]{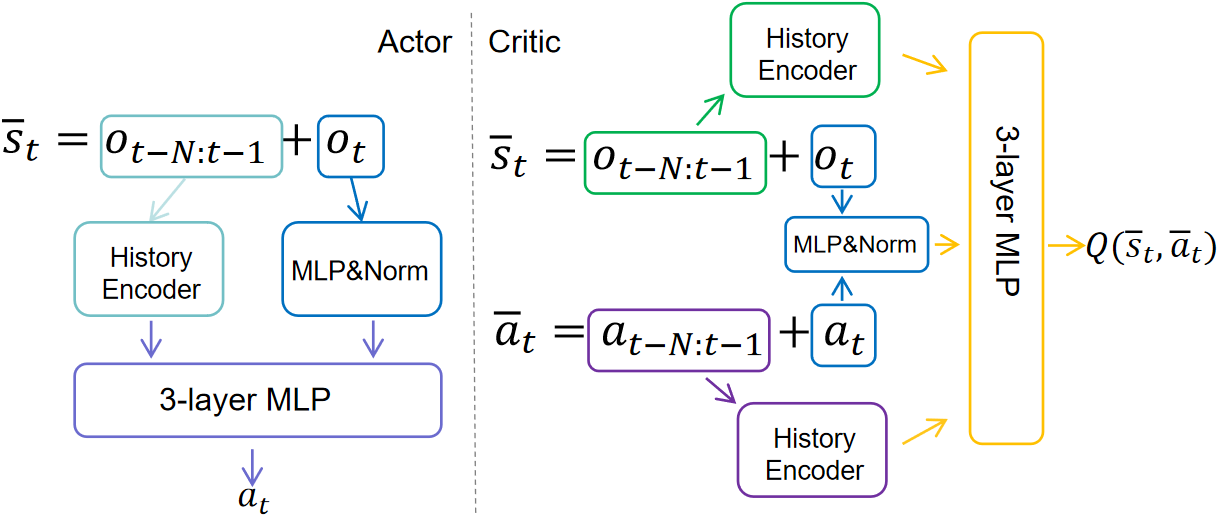}
\caption{Actor–critic architecture with history encoder for POMDP. Past observations and actions are processed by the history encoder to extract a compact representation. The current observation (or action) is normalized and concatenated with the encoded history. The resulting feature is then passed through a Multi-Layer Perceptron (MLP) to produce the output: an action for the actor (left) or a Q-value estimate for the critic (right).}
\label{fig:actor_critic}
\end{figure}

Recent advances in modeling historical information have been strongly driven by Transformer architectures with a self-attention mechanism \cite{vaswani2017attention}. Methods such as the Decision Transformer (DT) \cite{chen2021decision}, Online Decision Transformer (ODT) \cite{zheng2022online}, and Trajectory Transformer (TT) \cite{janner2021trajectory} demonstrate the remarkable capacity of self-attention to capture long-range temporal dependencies, achieving state-of-the-art performance in offline partially observable settings. However, full Transformer architectures typically require large-scale offline datasets, even for online learning, which introduces substantial computational overhead. These limitations highlight that while attention mechanisms are highly expressive for sequence modeling, deploying full Transformer architectures in RL remains challenging. This motivates exploring alternative ways to encode histories beyond purely Transformer-based models.

The Partially Observable Markov Decision Process (POMDP) \cite{astrom1965optimal} extends the standard Markov Decision Process (MDP) framework by introducing conditional observation probabilities, which map latent states to belief states that capture uncertainty arising from partial observability. While policies in MDPs typically map states to actions, POMDP policies are defined over belief states or histories of observations. Owing to its capacity to handle uncertainty, the POMDP framework has been widely adopted in various domains, including robotic control \cite{bhattacharya2021multiagent, dogar2020controlling, lee2021magic} and healthcare decision-making \cite{li2022optimizing, zhang2022diagnostic}.

POMDP is also addressed in reinforcement learning\cite{kaelbling1998planning}. Modern POMDP-based deep RL commonly employs Recurrent Neural Networks (RNNs) \cite{elman1990finding} to encode historical information for decision-making. Deep Recurrent Q-Learning \cite{Hausknecht2015DeepRQ} replaced the first fully connected layer with a Long Short-Term Memory (LSTM) \cite{hochreiter1997long} network to capture temporal dependencies, achieving significant performance gains over Deep Q-learning (DQN) \cite{mnih2015human} on Atari 2600 games. Song et al. \cite{Song2017RecurrentDP} combined RNNs with deterministic policy gradients to propose Recurrent DPG (RDPG), which was applied to partially observable obstacle-crossing tasks. Their results demonstrate that incorporating historical information effectively mitigates the challenges posed by partial observability. Ulrich et al. \cite{Ulrich2023Recurrent} integrated LSTM into the soft actor-critic framework, feeding the LSTM output along with observations into the agent, and achieved promising results in heat pump control. Meng et al. \cite{Meng2021Memory} investigated the impact of historical information under varying observation probabilities by integrating LSTM with Twin Delayed Deep Deterministic Policy Gradient (TD3) \cite{Fujimoto2018addressing}, claiming that LSTM can help increase performance in noisy observation and output-feedback control problems. Ni~\cite{Ni2022recurrent} built upon prior work and proposed a simple yet effective Recurrent Model-Free (RMF) framework tailored for a wide range of POMDP problems, which achieved state-of-the-art performance at the time. Based on this foundation, subsequent methods such as Ordinary Differential Equation RMF (ODERMF)\cite{Zhao2023ODE} and Context-Based Encoders \cite{luo2024efficient} further improved the performance of model-free RL under partial observability. However, these methods come with significantly higher computational complexity and require advanced Graphics Processing Units (GPUs). 

Motivated by the classical result of Bitmead \cite{bitmead1990mhe}, which demonstrates that the true state of an observable system can be inferred from a sequence of consecutive observations, we design a new history encoder that replaces recurrent networks with depthwise separable convolution and self-attention. We refer to this approach as the Convolution and Attention based Encoder (CAE), which can be incorporated into a TD3 algorithm to yield a new method, CAE-TD3.

On the algorithmic side, this design exploits the sequence modeling capability of attention while avoiding the substantial computational overhead of full Transformer architectures, resulting in a lightweight encoder that can be readily integrated into actor–critic methods. On the theoretical side, the parallel encoding mechanism of attention provides a principled way to approximate the transformation of a POMDP into an equivalent MDP under mild assumptions. Empirically, we demonstrate that CAE-TD3 achieves superior performance over recurrent baselines without increasing model complexity. Our main contributions are summarized as follows:

\begin{enumerate}
    \item  We propose a convolution and attention based history encoder that achieves superior performance over recurrent networks under the same parameter budget, demonstrating both efficiency and effectiveness in POMDP settings.
    \item We show that the parallel encoding offers a principled way to approximate POMDPs as MDPs under mild assumptions, opening a new direction for future research on partially observable reinforcement learning.
\end{enumerate}

\section{Preliminaries and Background}
\label{sec:pre}

This section introduces some preliminaries and background, including the basic problem formulation and the underlying techniques.

\subsection{Conversion from POMDP to MDP}
 \label{sec:pomdp}



A POMDP \cite{astrom1965optimal} extends the standard MDP framework \cite{Sutton2018Reinforcement} and is typically formulated as a 6-tuple $(\calS, \calA, \calP, \calR, \calS_o, \calO)$. Here, $\calS$ denotes the complete state space, $\calA$ the action space, $\calP:\calS\times\calA\times\calS \rightarrow \mathbb{R}$ the state transition probability function, and $\calR:\calS\times\calA \rightarrow \mathbb{R}$ the reward function. At each time step $t$, the agent selects an action $\bfa_t \in \calA$, causing the environment in the current state $\bfs_t \in \calS$ to transition to a new state $\bfs_{t+1}$ according to $\calP(\bfs_{t+1} \mid \bfs_t, \bfa_t)$, and the agent receives an immediate reward $\bfr_t = \calR(\bfs_t, \bfa_t)$.

In this scenario, the full state $\bfs_t \in \calS$ is not directly accessible. Instead, $\calS_o$ denotes the observation space, and $\calO: \calS \times \calA \times \calS_o \rightarrow \mathbb{R}$ is the observation function that defines the probability $\calO(\bfo_t \mid \bfs_t, \bfa_{t-1})$ of receiving observation $\bfo_t \in \calS_o$ given the state $\bfs_t$ and the previous action $\bfa_{t-1}$.

In this paper, we consider the simplified case where $\calO(\bfo_t \mid \bfs_t)$ is independent of the action. This setting includes a special case, covering all benchmark experiments in this study, where $\calS_o$ is a fixed subset of $\calS$, and the observation function $\calO$ is represented by a binary masking matrix containing only 0s and 1s, indicating a deterministic mapping from $\calS$ to $\calS_o$.




Let $\pi(\bfa_t \mid \bfs_t)$ be an action policy provided that $\bfs_t$ is accessible. Then the induced state transition function under the policy  is defined as:
\begin{align*}
\calP_\pi(\bfs_{t+1} \mid \bfs_t) =\sum_{\bfa_t } \calP(\bfs_{t+1} \mid \bfs_t,  \bfa_t)\pi(\bfa_t  \mid \bfs_t).    
\end{align*}
We now define a sequence of multi-step observation functions that relate observations at different time steps to the hidden state $\bfs_{t-N}$.
\begin{align*}
\calO_0(\bfo_{t-N} \mid \bfs_{t-N}) =& \calO(\bfo_{t-N} \mid \bfs_{t-N}) \\
\calO_1(\bfo_{t-N+1} \mid \bfs_{t-N}) =&\sum_{\bfs_{t-N+1}} \calO(\bfo_{t-N+1} \mid \bfs_{t-N+1})  \\ & \calP_\pi(\bfs_{t-N+1} \mid \bfs_{t-N}) \\
\vdots& \\
\calO_{N}(\bfo_t \mid \bfs_{t-N}) =&\sum_{\bfs_t,\cdots, \bfs_{t-N+1}} \calO(\bfo_t \mid \bfs_t)  \calP_\pi(\bfs_t \mid \bfs_{t-1})\\& \cdots \calP_\pi(\bfs_{t-N+1} \mid \bfs_{t-N}).
\end{align*}

Let the observation history be denoted compactly as:
\begin{align*}
\bar\bfs_t =\bfo_{t:t-N} := [\bfo_t, \bfo_{t-1}, \ldots, \bfo_{t-N}].    
\end{align*}
The same vector-concatenation notation is used throughout this paper.
We can then define a joint observation function  that maps this sequence to the latent state $\bfs_{t-N}$:
\begin{align*}
\bar{\calO}(\bar\bfs_t \mid \bfs_{t-N}) := \prod_{k=0}^{N} \calO_k(\bfo_{t-N+k} \mid \bfs_{t-N}).
\end{align*}
Assuming the system is observable (i.e., the observation sequence is informative about the latent state, without requiring knowledge of its prior distribution), we define the posterior belief over the initial hidden state $\bfs_{t-N}$ given the observation history $\bar\bfs_t$ as $\calQ(\bfs_{t-N} \mid \bar\bfs_t)$. 

From this posterior, we recursively estimate the subsequent hidden states using the transition model:
\begin{align*}
\calQ_0(\bfs_{t-N} \mid \bar\bfs_t) =& \calQ(\bfs_{t-N} \mid \bar\bfs_t) \\
\calQ_1(\bfs_{t-N+1} \mid \bar\bfs_t) =& \sum_{\bfs_{t-N+1}} \calP_\pi(\bfs_{t-N+1} \mid \bfs_{t-N}) \\& \calQ_0(\bfs_{t-N} \mid \bar\bfs_t) \\
\vdots & \\
\calQ_{N}(\bfs_t \mid \bar\bfs_t) =& \sum_{\bfs_{t-1}} \calP_\pi(\bfs_t \mid \bfs_{t-1}) \calQ_{N-1}(\bfs_{t-1} \mid \bar\bfs_t)
\end{align*}

The observation history evolves deterministically  as a sliding window of fixed length:
\begin{align*}
\bar\bfs_{t+1} = \bfo_{t+1:t-N+1} = 
 \bfo_{t+1}  \begin{bmatrix}
1 & 0_{1\times N}
\end{bmatrix} 
+ \bar\bfs_t \begin{bmatrix}
0_{N\times 1} & I_{N \times N} \\
0 & 0_{1 \times N}
\end{bmatrix} .
\end{align*}
This implies that the transition probability of $\bar\bfs_{t+1}$, given $\bar\bfs_t$ and action $\bfa_t$, is determined entirely by the probability distribution over the next observation $\bfo_{t+1}$. Hence:
\begin{align*}
&\bar{\calP}_\pi(\bar\bfs_{t+1} \mid \bar\bfs_t, \bfa_t) = \bar{\calP}_\pi(\bfo_{t+1} \mid \bar\bfs_t, \bfa_t) \\
=& \sum_{\bfs_{t+1}, \bfs_t} \calO(\bfo_{t+1} \mid \bfs_{t+1}) \, \calP(\bfs_{t+1} \mid \bfs_t, \bfa_t) \, \calQ_{N}(\bfs_t \mid \bar\bfs_t).
\end{align*}
Note that $\calO$ and $\calQ_{N}$ depend on $\pi$.

For a given observation $\bar\bfs_t$, the distribution of $\bfs_t $ is $\calQ_{N}(\bfs_t \mid \bar\bfs_t)$.
Under the control policy $\pi(\bfa_t  \mid \bfs_t)$,   the distribution of $\bfa_t$ given this observation is 
\begin{align*}
\varpi(\bfa_t \mid \bar\bfs_t) = \sum_{\bfs_t} \pi(\bfa_t \mid \bfs_t) \cdot \calQ_{N}(\bfs_t \mid \bar\bfs_t).
\end{align*}
 
In summary, the problem is reformulated as a new MDP $(\bar \calS, \calA, \bar{\calP}_\pi, \bar\calR)$ where $\bar\calS=\calS_o\times\cdots\times\calS_o$ denotes the state space, $\calA$ the action space, $\bar{\calP}_\pi:\bar\calS\times\calA\times\bar\calS \rightarrow \mathbb{R}$ the state transition probability function, and $\bar\calR:\bar\calS\times\calA \times  \mathbb{R}\rightarrow \mathbb{R}$ the reward distribution. At each time step $t$, the agent selects an action $\bfa_t \in \calA$, causing the environment in the current state $\bar\bfs_t \in \bar\calS$ to transition to a new state $\bar\bfs_{t+1}$ according to $\bar{\calP}_\pi(\bar\bfs_{t+1} \mid \bar\bfs_t, \bfa_t)$, and the agent receives an immediate reward $\bfr_t =\calR(\bfs_t, \bfa_t)\sim \bar\calR(\cdot \mid \bar\bfs_t, \bfa_t)$.

For this new MDP,  the goal is to learn an action policy $\varpi(\bfa_t \mid \bar\bfs_t)$ based on the fully observable state $\bar\bfs_t$. 
This formulation allows a RL framework to be applied, despite two key challenges.
The first challenge is that the transition function $\bar{\calP}_\pi$ depends on the action policy itself. While model-free RL methods do not require explicit knowledge of $\bar{\calP}_\pi$, facilitating learning in this setting, both the model and the policy must be learned in a mutually dependent manner.

The second challenge arises from the complexity of the new state $\bar\bfs_t$, which is typically much higher than that of the original state $\bfs_t$, especially when $\bar\bfs_t$ is constructed to retain sufficient information from the observation history. This increased complexity requires specialized techniques to process $\bar\bfs_t$ before a conventional RL policy can be effectively applied.

\subsection{Observation History}

The core contribution of this work is a novel history encoder for processing historical observations, which integrates two key operations: depthwise separable convolution and multi-head attention. The background of these operations is briefly reviewed below.

Depthwise separable convolution \cite{Chollet2016XceptionDL} is a streamlined variant of standard convolution that reduces both parameter count and computational cost. It decomposes the operation into two steps: (1) a depthwise convolution, which applies a single filter to each input channel, and (2) a pointwise convolution, which uses a $1 \times 1$ convolution to linearly combine the outputs of the depthwise stage.

This factorization enables efficient extraction of spatial and cross-channel features while maintaining strong representational capacity. It has been widely adopted in lightweight neural architectures such as MobileNet \cite{Sinha2019mobile} and is particularly well suited to real-time and resource-constrained applications.


Multi-head attention \cite{vaswani2017attention} is a fundamental component of Transformer architectures, enabling the model to capture diverse patterns across different representation subspaces. Given queries $Q$, keys $K$, and values $V$, the attention mechanism is defined as:
\begin{equation*}
    \text{Attention}(Q, K, V) = \text{softmax}\left(\frac{QK^\top}{\sqrt{d_k}}\right)V,
\end{equation*}
where $d_k$ is the dimensionality of the keys. In multi-head attention, this operation is executed in parallel across multiple heads, each using distinct learned projections of $Q$, $K$, and $V$. The outputs from all heads are then concatenated and linearly transformed.

This structure allows the model to attend to information from multiple perspectives simultaneously, making it particularly effective at capturing long-range dependencies. In this work, we employ self-multi-head attention, where $Q$, $K$, and $V$ are all derived from the same input matrix.

\begin{figure*}[hbtp]
\centering
\includegraphics[width=0.8\linewidth]{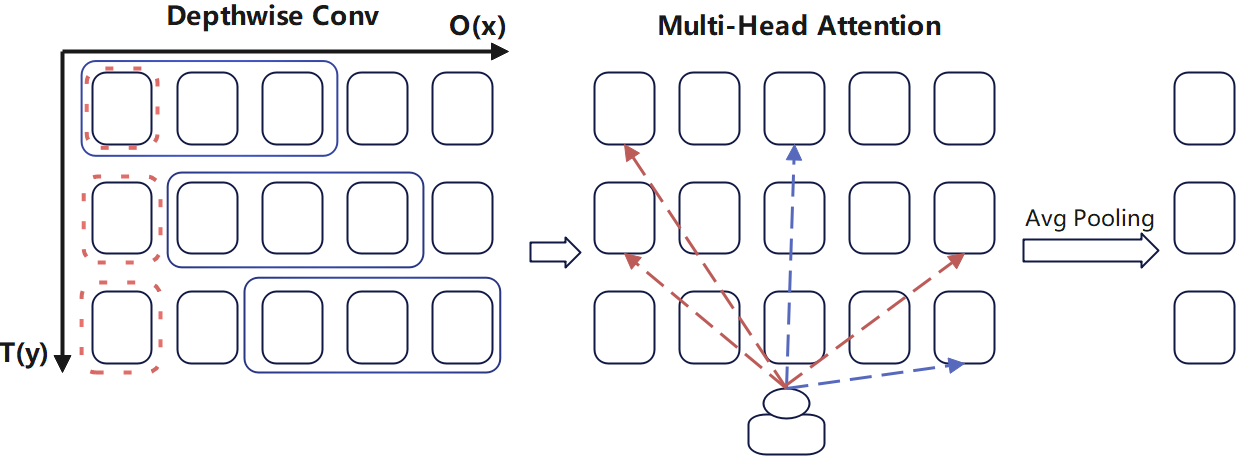}
\caption{Procedure of the history encoder. The white blocks represent observation elements. Depthwise convolution merges information along both the observation dimension O(x) and the time dimension T(y). Multi-head self-attention then captures global dependencies, and average pooling compresses the result into a compact hidden representation.}
\label{fig:history encoder}
\end{figure*}

A relevant technique is the LSTM network \cite{hochreiter1997long}, a class of RNNs designed to capture long-range temporal dependencies in sequential data.
Unlike vanilla RNNs, which suffer from vanishing and exploding gradients, LSTM introduces gating mechanisms, including input, output, and forget gates, to regulate the flow of information and preserve memory over time.

In RL, LSTMs are widely employed to encode historical observation–action sequences, enabling agents to make more informed decisions under partial observability. At each time step, the LSTM processes the input (e.g., the current observation or observation–action pair) and updates its hidden state, which serves as a compact representation of past information. This recurrence provides the policy or value function with temporal context that extends beyond the current input.

\subsection{TD3 Algorithm}

The history encoder reconstructs full state information from the observation history and is integrated into the RL policy. We adopt the TD3 algorithm as our baseline, which we revisit below. TD3 \cite{Fujimoto2018addressing} is an actor-critic algorithm designed for continuous control tasks. It extends the Deep Deterministic Policy Gradient (DDPG) framework by addressing overestimation bias in value estimation, which is a common issue in actor-critic methods.

TD3 introduces three key modifications: (i) the use of two critic networks to compute the minimum Q-value for target updates, (ii) delayed policy updates to improve stability, and (iii) target policy smoothing by adding clipped noise to the target action. Specifically, the target Q-value is computed as:
\begin{equation*}
\mathbf{y} =\mathbf{ r} + \gamma \min_{i=1,2} Q_{\theta_i'}(\bfs', \pi_{\phi'}(\bfs') + \epsilon),    
\end{equation*}
where $\epsilon \sim \mathcal{N}(0, \sigma)$ is clipped noise, $\theta_1'$ and $\theta_2'$ are target critic parameters, and $\phi'$ denotes the target actor.

\section{Methodology}

We first present the innovative structure of the history encoder and explain how it is integrated into the actor–critic framework. Specifically, we introduce a new RL algorithm built upon TD3 and its variants.

\subsection{History Encoder}

The history encoder comprises two main components: depthwise separable convolution and multi-head self-attention, collectively referred to as the CAE. As illustrated in Figure~\ref{fig:history encoder}, the depthwise convolution is applied independently along the observation dimension (blue boxes, horizontal) and the temporal dimension (brown boxes, vertical). This design, compared with conventional CNNs, reduces coupling between heterogeneous features across the temporal and observation axes, thereby better aligning with the structural characteristics of our input data.

However, since depthwise convolution treats all temporal and observational elements uniformly, we further incorporate a multi-head self-attention mechanism (center) to assign adaptive importance to different positions across both axes. The attention mechanism captures global dependencies, as highlighted by brown and blue arrows, enabling the model to focus on strongly correlated temporal or spatial features.

Finally, the attention output is aggregated via average pooling into a compact one-dimensional hidden representation, which serves as the encoded summary of historical information. 

\subsection{Actor-Critic Framework}
\label{sec: align with OPMDP}

This section explains how the proposed CAE is embedded within an actor–critic framework. In standard actor–critic algorithms, the actor generates an action based on the current observation, while the critic evaluates the quality of that action by estimating its Q-function. 

As defined in Section~\ref{sec:pomdp}, the agent's observation is $\bar{\bfs}_t$, and the action at time $t$ is $\bfa_t$. Accordingly, the actor network operates as follows:
\begin{align}
    \bfa_t &= \pi_\phi(\bar{\bfs}_t),  \label{eq:actor}
\end{align}
The function $\pi_\phi$ represents the mapping from $\bar{\bfs}_t$ to $\bfa_t$, as illustrated in Figure~\ref{fig:actor_critic}. This mapping comprises the history encoder, the MLP, and the normalization module.

To account for the state $\bar{\bfs}_t$, which is reconstructed from historical observations, we also define $\bar{\bfa}_{t} = \bfa_{t-N:t}$ as the concatenation of past actions. This concatenated action sequence is further incorporated into the critic operation as follows:
\begin{align}
 &Q_\theta(\bar{\bfs}_t, \mathbf{\bar{a}}_t) \nonumber \\
 =& \mathbb{E}_{\bar{\bfs}_{t+1},\bfr_t} \left[ \bfr_t + \gamma Q_{\theta'}(\bar{\bfs}_{t+1}, [\bfa_{t-N+1:t}, \pi_{\phi'}(\bar{\bfs}_{t+1})]) \right]. \label{eq:critic}
\end{align}
Similarly, the function $Q_\theta$ represents the mapping illustrated in Figure~\ref{fig:actor_critic}, which consists of two history encoders, an MLP, and a normalization module. Notably, the three history encoders in Figure~\ref{fig:actor_critic} are independent of one another, as they are designed to capture different information. To ensure numerical stability, we apply the same pooling-based normalization method in both the actor and the critic, consistent with the design shown in Figure~\ref{fig:history encoder}.

Since the framework involves two pairs of actor and critic networks, 
$\phi$ and $\phi'$ denote the actor and target actor parameters, 
while $\theta$ and $\theta'$ denote the critic and target critic parameters, respectively.
 
The corresponding training objectives for the actor and critic networks are defined as follows:
\begin{align}
    J(\phi) &= - Q_\theta(\bar{\bfs}_t,[\bfa_{t-N:t-1}, \pi_\phi(\bar{\bfs}_t)]), \label{eq:actor train}\\
    J(\theta)&= \bigg(\bfr_t + \gamma Q_{\theta'}(\bar{\bfs}_{t+1}, [\bfa_{t-N+1:t}, \pi_{\phi'}(\bar{\bfs}_{t+1}) \nonumber\\ & -Q_{\theta}(\bar{\bfs}_t, \mathbf{\bar{a}}_t)\bigg)^2.\label{eq:critic train}
\end{align}
The empirical expectations of these objectives are computed during training using samples drawn from the replay buffer, which has the structure $(\bar{\bfs}_t, \bar{\bfa}_t, \bfr_{t}, \bar{\bfs}_{t+1})$.  
Building on the TD3 algorithm as the baseline, the complete algorithm incorporating the CAE, referred to as \texttt{CAE-TD3}, is summarized in Algorithm~\ref{alg:datd3}. It is noted that the updates of the target network weights, $\phi'$ and $\theta'$, from $\phi$ and $\theta$ follow the standard TD3 procedure and are therefore not repeated in this algorithm.



\begin{algorithm}
 \caption{CAE-TD3}
 \label{alg:datd3}
 \begin{algorithmic}[1]
 \renewcommand{\algorithmicrequire}{\textbf{Input:}}
 \renewcommand{\algorithmicensure}{\textbf{Output:}}
 \REQUIRE Environment with observation space $\calS_o$ and action space $\calA$
 \ENSURE  Learned policy $\pi$ and Q-value function $Q$
 
  \STATE Initialize policy network $ \pi_{\phi}$ and Q-network $Q_{\theta}$ with random weights. Initialize target networks $\pi'$ and $Q'$ with weights  $\phi' \leftarrow \phi$ and $\theta' \leftarrow \theta$.

  \STATE  Initialize two empty queues $\mathbf{H_s}$ and $\mathbf{H_a}$ with a fixed length $\mathbf{N+1}$.

  \FOR {$episode = 1$ to $M$}
        \STATE Collect history observation sequence  $\mathbf{H_s} =\bfo_{0:N}$ to form $\bar{\bfs}_0$ and history action sequence $\mathbf{H_a} = a_{0:N-1}$ from random exploration.
        
        \WHILE{Not Terminated}
            \STATE  Select action $\bfa_t = \pi_\phi(\bar{\bfs}_t) + \mathcal{N}_t$ according to the current policy and exploration noise.
            \STATE Execute $\bfa_t$ in the environment and observe next observation $\bfo_{t+1}$ and reward $\bfr_{t}$.
            \STATE Append $\bfo_{t+1}$ to $\mathbf{H_s}$ and $\bfa_t$ to $\mathbf{H_a}$ to form $\bar{\bfs}_{t+1}$ and $\bar{\bfa}_t$, respectively.
            \STATE Store transition $(\bar{\bfs}_t, \bar{\bfa}_t, \bfr_{t}, \bar{\bfs}_{t+1})$ in replay buffer $\mathcal{B}$
            \STATE Update the network wights $\phi$ and $\theta$ according to the objectives in \eqref{eq:actor train} and \eqref{eq:critic train}.
  \ENDWHILE
  \ENDFOR   
 \end{algorithmic} 
 \end{algorithm}

\subsection{CAE-TD3 Variants}

\label{sec:datd3 full obs}

\begin{figure}[ht]
\centering
\includegraphics[width=0.9\linewidth]{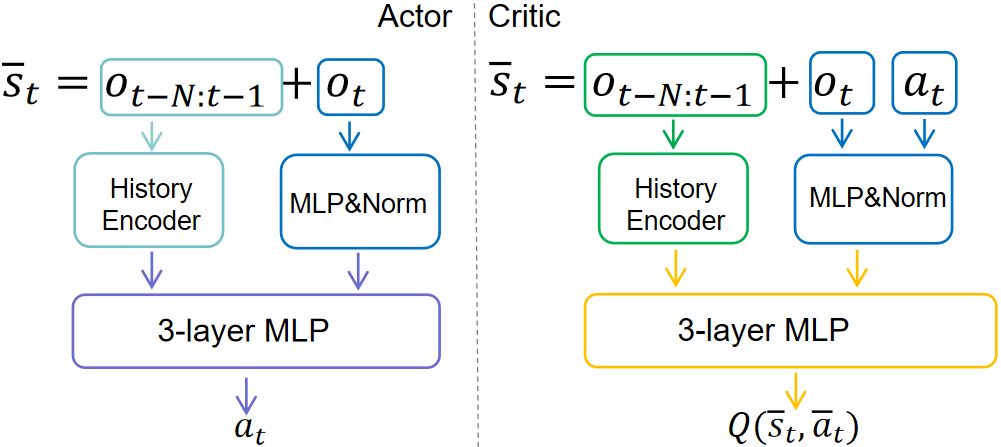}
\caption{Actor–critic architecture of CAE-TD3 with full observation (CAE-TD3-FO).}
\label{fig: method4 network}
\end{figure}

For a fully observable MDP with $\mathcal{S} = \mathcal{S}_o$, an RL algorithm does not require historical information. However, both empirical studies and algorithmic advancements \cite{zheng2022online, Meng2021Memory, Ulrich2023Recurrent} have shown that incorporating redundant historical information can enhance performance by yielding more stable and expressive representations. In the fully observable setting, the proposed \texttt{CAE-TD3} algorithm retains observation history to introduce structured redundancy while discarding historical action memory. Figure~\ref{fig: method4 network} illustrates the network architecture of \texttt{CAE-TD3} under the full-state observation scenario, hereafter abbreviated as \texttt{CAE-TD3-FO}.

In addition, we explored several architectural variants of \texttt{CAE-TD3} during the design process. The final structure was selected based on empirical results, which demonstrated its superior performance. A detailed comparison and analysis of alternative designs is provided in Appendix~\ref{sec:variants}, highlighting the limitations of those variants and further justifying the effectiveness of the proposed architecture.

\section{Experimental Results}
\label{sec:exp}

This section reports a comparison of experimental results across various environments and algorithms to evaluate the effectiveness of \texttt{CAE-TD3}.

\subsection{Environments and Baseline Algorithms}

The experiments are conducted on the Gymnasium benchmark \cite{Towers2023Gym}. Following prior work~\cite{Meng2021Memory,Ni2022recurrent}, we select four representative continuous-control environments: \texttt{Ant-v4}, \texttt{Inverted Pendulum-v4} (referred to as \texttt{Pendulum-v4}), \texttt{Hopper-v4}, and \texttt{Walker-v4}. Each environment’s observations consist of position and velocity sensor readings. In the partial-observation experiments, we retain only the position components. The corresponding observation dimensions under each setting are summarized in Table~\ref{tab:obs detail}.

\begin{table}[ht]
    \centering
    \begin{tabular}{lcccc}
        \toprule
        & Ant-v4 & Pendulum-v4 & Hopper-v4 & Walker-v4 \\
        \midrule
        Position & 13 & 2 & 5 & 8 \\
        Velocity & 14 & 2 & 6 & 9 \\
        \bottomrule
    \end{tabular}
    \caption{Dimensions of the position and velocity state spaces in four environments. In the partial-observation experiments, only position sensor data are used.}
    \label{tab:obs detail}
\end{table}

We select two baselines for comparison: Memory-Based LSTM-TD3 \cite{Meng2021Memory} (\texttt{LSTMTD3}) and Recurrent Model-Free RL \cite{Ni2022recurrent} (\texttt{RMF}). Since the original TD3 algorithm does not incorporate any mechanism for temporal representation or memory, we also include a modified variant called Fixed-Window TD3 (\texttt{FWTD3}). In \texttt{FWTD3}, inputs from multiple consecutive time steps are concatenated to simulate temporal memory, allowing the agent to leverage a limited historical context. For \texttt{LSTMTD3} and \texttt{RMF}, we adopt the default hyperparameters reported in the respective papers, while \texttt{FWTD3} uses the same hyperparameters as TD3 with a fixed window length of three.

\subsection{Partially Observable Environments}

\begin{figure*}[hbtp]
\centering
\includegraphics[width=0.95\linewidth]{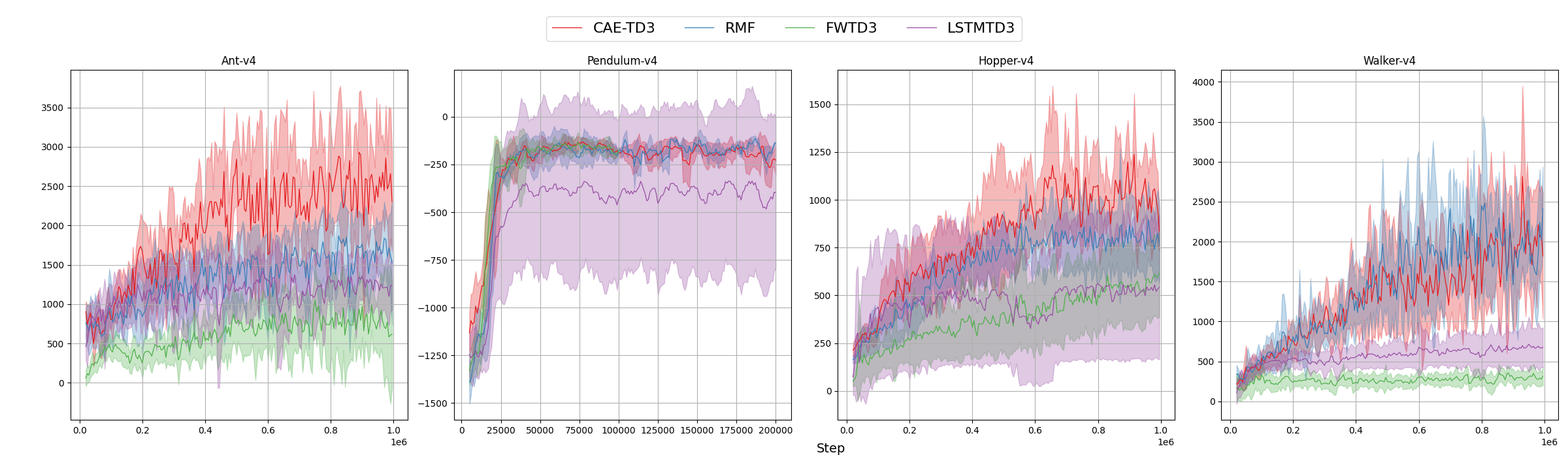}
 \caption{Training performance across four partially observable environments comparing CAE-TD3 with baseline algorithms.}
\label{fig:mainresult}
\end{figure*}

We trained agents in each environment for 1,000,000 steps (100,000 steps for \texttt{Pendulum-v4}). The performance of \texttt{CAE-TD3} and the baseline algorithms across four partially observable environments is shown in Figure~\ref{fig:mainresult}. The corresponding results in Table~\ref{tab:combined_results} are reported as the mean $\pm$ one standard deviation, calculated over the final 200,000 training steps for each random seed.

\begin{table*}[t]
\centering
\caption{Average return and standard deviation across four environments under both partial and full observability. Bold values indicate the best result in each column.}
\label{tab:combined_results}
\begin{tabular}{lcccc}
\toprule
&  Ant-v4 & Pendulum-v4 & Hopper-v4 & Walker-v4 \\
\midrule
\multicolumn{5}{c}{{Partial Observability}} \\
\midrule
CAE-TD3 (Ours) & $\mathbf{2520.99 \pm 317.59}$ & $\mathbf{-157.70 \pm 11.77}$ & $\mathbf{981.17 \pm 66.70}$ & $1907.39 \pm 168.27$ \\
RMF & $1676.61 \pm 171.34$ & $-178.69 \pm 22.1$ & $802.22 \pm 48.14$ & $\mathbf{1970.68 \pm 179.72}$ \\
FWTD3 & $725.27 \pm 165.50$ & $-160.70 \pm 10.44$ & $591.22 \pm 16.10$ & $303.45 \pm 23.34$ \\
LSTMTD3 & $1203.98 \pm 98.37$ & $-391.33 \pm 31.56$ & $536.28 \pm 17.10$ & $679.70 \pm 35.53$ \\
\midrule
\multicolumn{5}{c}{{Full Observability}} \\
\midrule
CAE-TD3-FO (Ours)& $\mathbf{4586.06 \pm 97.39}$ & $-187.78 \pm 94.27$ & $2302.01 \pm 104.85$ & $\mathbf{3903.26 \pm 210.47}$ \\
TD3 & $3239.46 \pm 219.17$ & $\mathbf{-154.79 \pm 7.34}$ & $3073.67 \pm 111.13$ & $3326.81 \pm 265.54$ \\
LSTMTD3 & $3856.28 \pm 106.30$ & $-146.91 \pm 7.82$ & $\mathbf{3136.92 \pm 112.05}$ & $3383.26 \pm 125.47$ \\
\bottomrule
\end{tabular}
\end{table*}

In \texttt{Ant-v4}, \texttt{CAE-TD3} consistently outperforms all baselines in both final return and sample efficiency. The margin is particularly pronounced against \texttt{FWTD3}, underscoring the necessity of structured processing over historical concatenation. 

In \texttt{Pendulum-v4}, \texttt{CAE-TD3} achieves the highest return, followed by \texttt{RMF} and \texttt{FWTD3}. In contrast, \texttt{LSTMTD3} performs significantly worse, suggesting that pure recurrent modeling without feature engineering leads to instability in learning.

In \texttt{Hopper-v4}, \texttt{CAE-TD3} shows consistent improvement over time and achieves the highest average return among all methods. \texttt{RMF} again performs competitively, though slightly worse, while \texttt{LSTMTD3} suffers from lower asymptotic performance. \texttt{FWTD3} remains the weakest, struggling to cope with the task’s complexity.

In \texttt{Walker-v4}, a more complex environment, \texttt{CAE-TD3} and \texttt{RMF} reach comparable peak performance, though \texttt{RMF} exhibits slightly larger variance. In contrast, \texttt{LSTMTD3} and \texttt{FWTD3} perform markedly worse, highlighting their limited ability to capture intricate temporal dependencies.


Overall, \texttt{FWTD3} performs consistently worse across all tasks except for the simple environment \texttt{Pendulum-v4}, supporting the motivation that merely concatenating past observations fails to provide effective temporal abstraction. Without explicit encoding mechanisms, the model is unable to extract useful representations from raw historical sequences.

\texttt{CAE-TD3} achieves the best or near-best performance in all four environments, owing to its architectural design: historical information is encoded through depthwise separable convolutions and multi-head attention. This structure enables the model to treat different time steps uniformly while flexibly attending to the most relevant information. 

By contrast, LSTM-based models such as \texttt{LSTMTD3} inherently bias toward recent inputs due to their recurrent nature, which constrains their capacity for long-range representation. \texttt{RMF} outperforms \texttt{LSTMTD3}, a result we attribute to its architectural separation of historical and current information streams. This distinction decouples gradient flows for different functions, reduces input-level interference, and promotes clearer functional specialization.

Taken together, these results demonstrate that \texttt{CAE-TD3} strikes an effective balance between memory utilization and architectural modularity. Its encoding design offers a structured and learnable approach to leveraging historical information, yielding robust performance in both simple and complex environments.

\subsection{Fully Observable Environments}

\begin{figure*}[hbtp]
\centering
\includegraphics[width=0.95\linewidth]{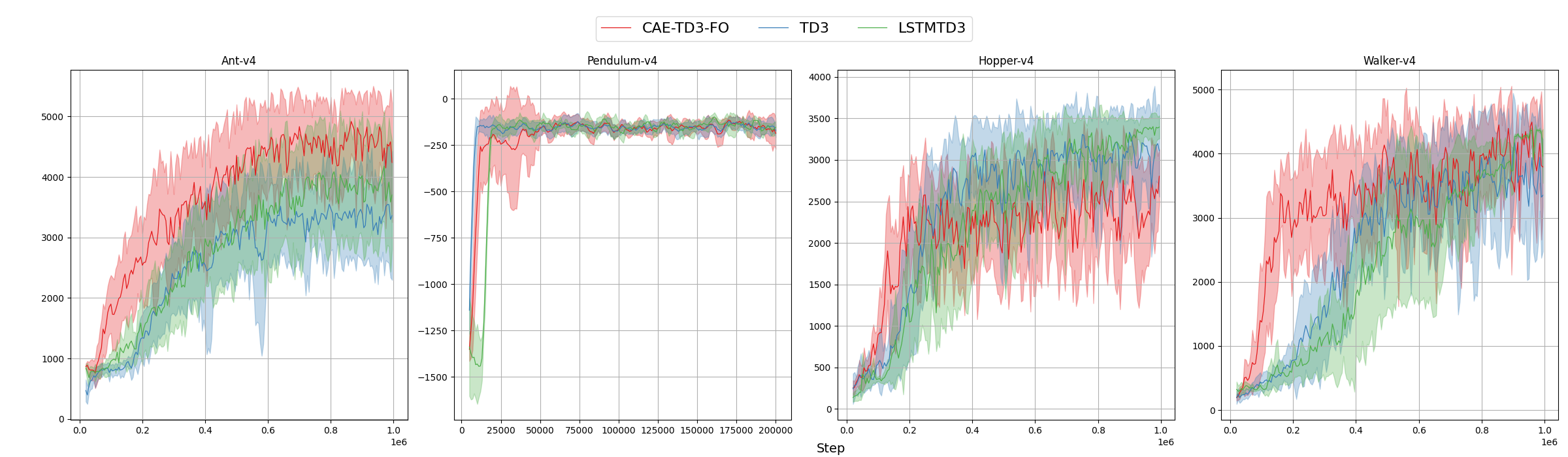}
\caption{Training performance across four fully observable environments comparing CAE-TD3-FO with baseline algorithms.}
\label{fig:complete obs res}
\end{figure*}

We further evaluate the performance of \texttt{CAE-TD3} in fully observable environments. In this setting, the network architecture follows the design outlined in Section~\ref{sec:datd3 full obs} and is denoted as \texttt{CAE-TD3-FO}, where only historical observations are retained while action history is excluded. Since \texttt{RMF} is tailored for partially observable environments, it is omitted here. Likewise, \texttt{FWTD3} reduces to the original \texttt{TD3}. Accordingly, we compare \texttt{CAE-TD3-FO} against \texttt{TD3} and \texttt{LSTMTD3} under fully observable conditions. The training curves are presented in Figure~\ref{fig:complete obs res}, and the corresponding statistical results over the final 200,000 training steps are summarized in Table~\ref{tab:combined_results}.
 
\texttt{CAE-TD3-FO} remains competitive and frequently outperforms both \texttt{TD3} and \texttt{LSTMTD3}. These results suggest that incorporating structured observation history, even in fully observable settings, improves learning stability and sample efficiency. By contrast, \texttt{LSTMTD3} demonstrates higher variance and slower convergence, underscoring the advantage of \texttt{CAE-TD3}’s lightweight history encoder over recurrent architectures.

\subsection{Effect of History Length}

\begin{figure}[ht]
\centering
\includegraphics[width=0.95\linewidth]{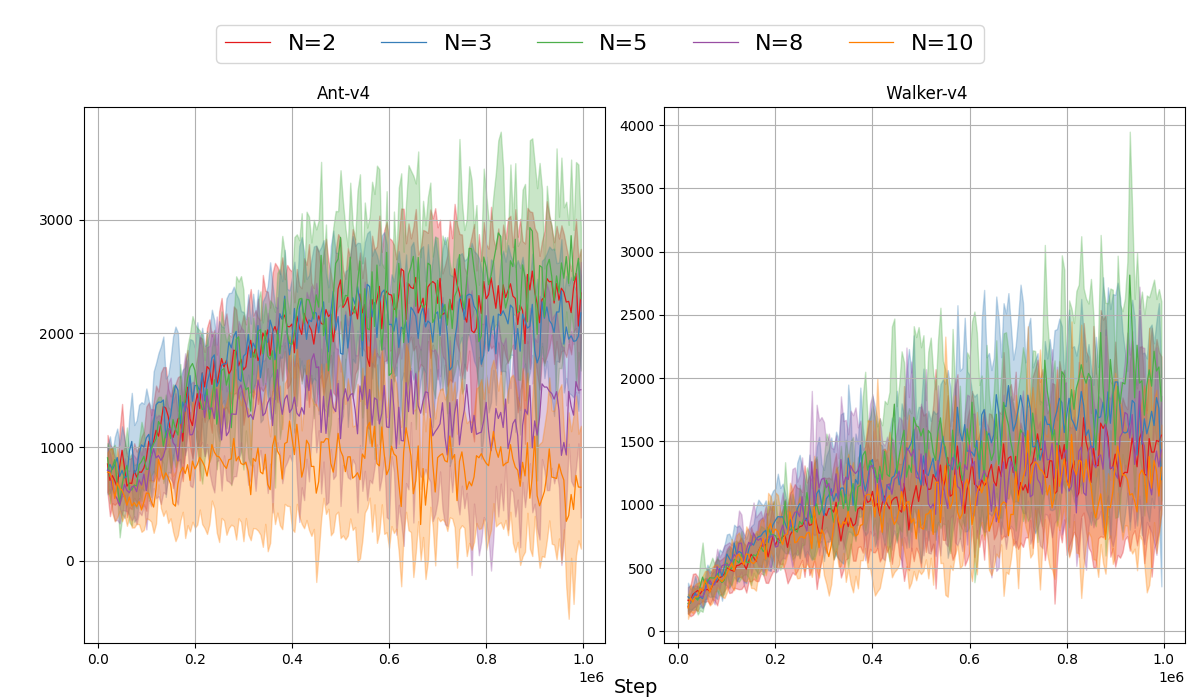}
\caption{Training performance with different history lengths in \texttt{Ant-v4} and \texttt{Walker-v4} environments.}
\label{fig:history length}
\end{figure}

Since historical observations play a central role in the proposed algorithm, it is important to examine the effect of history length $N$. We therefore conduct an ablation study in the \texttt{Ant-v4} and \texttt{Walker-v4} environments. The results, presented in Figure~\ref{fig:history length}, are smoothed using a moving average with a window size of 5 for clarity.

The findings suggest that \texttt{Ant-v4} benefits more from shorter history lengths, with the best performance obtained at a window size of 3. In contrast, \texttt{Walker-v4} shows relatively minor differences across history lengths during the early training stages. However, as training progresses, models with history lengths of 3 and 5 exhibit clear advantages, with the length-5 setting ultimately yielding the highest performance. These results indicate that appropriately longer observation sequences can be particularly beneficial in more complex environments with delayed dynamics.

\subsection{Computational Cost}

We present a comparison of the computational overhead associated with different methods. All experiments were conducted on a compute node of the Australian National Computational Infrastructure (NCI), equipped with a single NVIDIA Tesla V100 GPU and an Intel Xeon Gold 6148 CPU.

The detailed results are summarized in Table~\ref{tab:compu_cost}. Our comparison focuses on the number of parameters in the history encoder component (not applicable for \texttt{FWTD3}). For \texttt{CAE-TD3}, we report the total parameter count across the three history encoders, together with the two accompanying MLP and normalization layers. Because the number of parameters and runtime can vary depending on the environment (e.g., observation and action dimensions), we use the \texttt{Hopper-v4} environment as a representative example for analysis. The results show that the proposed \texttt{CAE-TD3} does not incur substantially higher computational cost than other encoding-based algorithms, and this trend holds across environments.

\begin{table}[t]
\centering
\caption{Comparison of model parameter counts and training time.}
\label{tab:compu_cost}
\begin{tabular}{lcc}
\toprule
  & Parameters & Training Time (hours)\\
\midrule
CAE-TD3 (Ours)    & 200,231   & 6.48 \\
RMF      & 199,104   & 7.78 \\
LSTMTD3  & 150,658   & 4.92 \\
FWTD3    & NA   & 1.83 \\
\bottomrule
\end{tabular}
\end{table}

\section{Conclusion}
\label{sec:conclusion}

In this work, we proposed a history encoder for processing historical observations in RL under POMDPs. Building on this encoder, we developed \texttt{CAE-TD3}, a history-aware actor–critic algorithm. By structuring observation histories through convolution and attention, \texttt{CAE-TD3} achieves strong performance under both partial and full observability. Extensive experiments demonstrate its effectiveness across diverse continuous control tasks.  Future research could explore scaling the encoder to high-dimensional sensory inputs, extending the approach to multi-agent and hierarchical settings, and investigating adaptive history lengths for greater efficiency. These directions may further broaden the applicability of lightweight temporal encoding and reinforce its role in advancing robust and scalable decision-making.

\section{Appendix}
\label{sec:appendix}

\label{sec:variants}

Variants of \texttt{CAE-TD3}, while preserving its core mechanisms, are examined in this section. The architectures of the different variants are illustrated in comparison with the baseline design shown in Figure~\ref{fig:actor_critic}. In all cases, the actor and critic remain fully separated, without sharing any network modules. The performance of these variants is then evaluated across selected environments.

\begin{figure}[ht]
\centering
\includegraphics[width=0.9\linewidth]{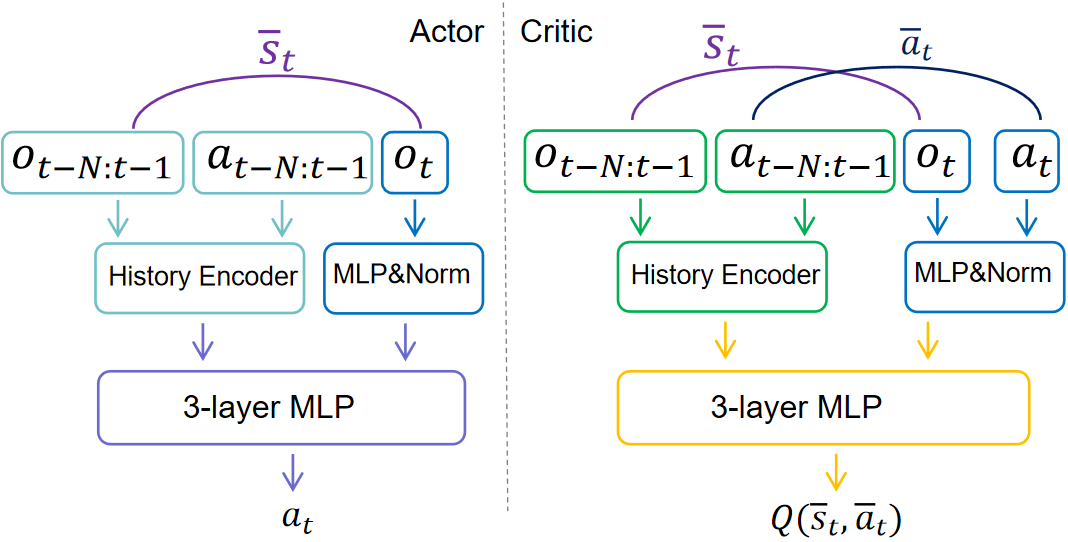}
\caption{First variant of CAE-TD3, where actions and observations are jointly processed.}
\label{fig: method1}
\centering
\includegraphics[width=0.9\linewidth]{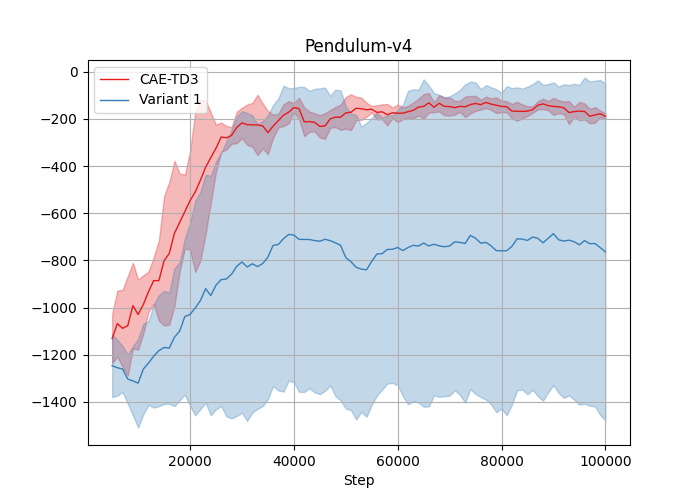}
\caption{Comparison of training performance between CAE-TD3 and Variant~1 in Pendulum-v4.}
\label{fig: cmp inc pendulum method 1}
\end{figure}

\texttt{Variant~1}: As illustrated in Figure~\ref{fig: method1}, historical observations and actions are concatenated and jointly processed by a shared history encoder, which is used by both the actor and the critic. This design aims to exploit convolution and attention mechanisms to capture transition dynamics directly from raw observation–action sequences, thereby facilitating training. However, the experimental results in Figure~\ref{fig: cmp inc pendulum method 1} show that such joint processing of observations and actions is largely ineffective.

\begin{figure}[ht]
\centering
\includegraphics[width=0.9\linewidth]{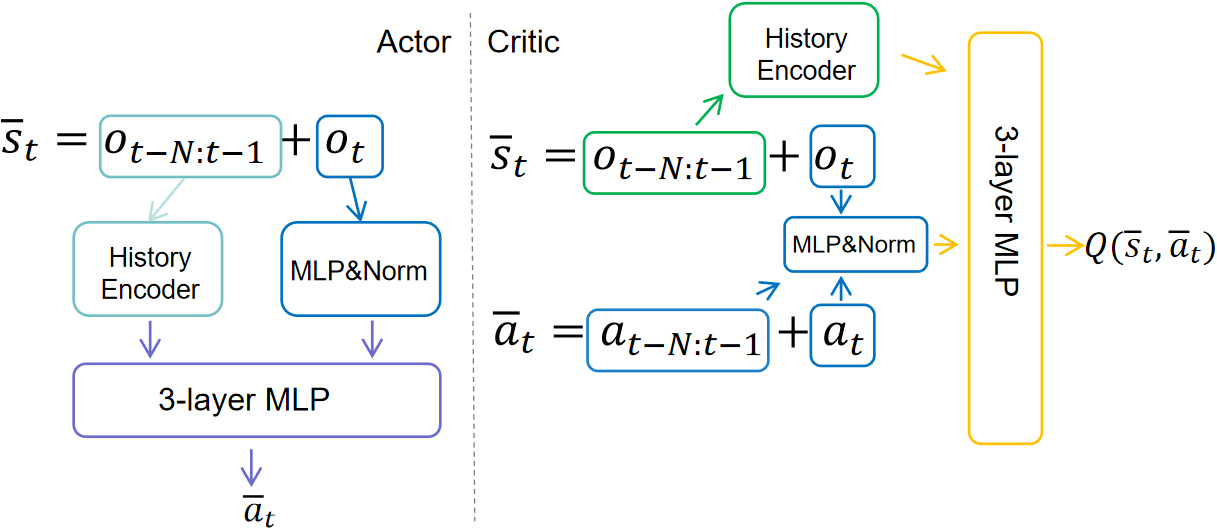}
\caption{Second variant of CAE-TD3, where the actor generates the complete $\bar{\bfa}_t$ and the critic evaluates $(\bar{\bfs}_t,\bar{\bfa}_t)$.}
\label{fig: method2}
\centering
\includegraphics[width=0.95\linewidth]{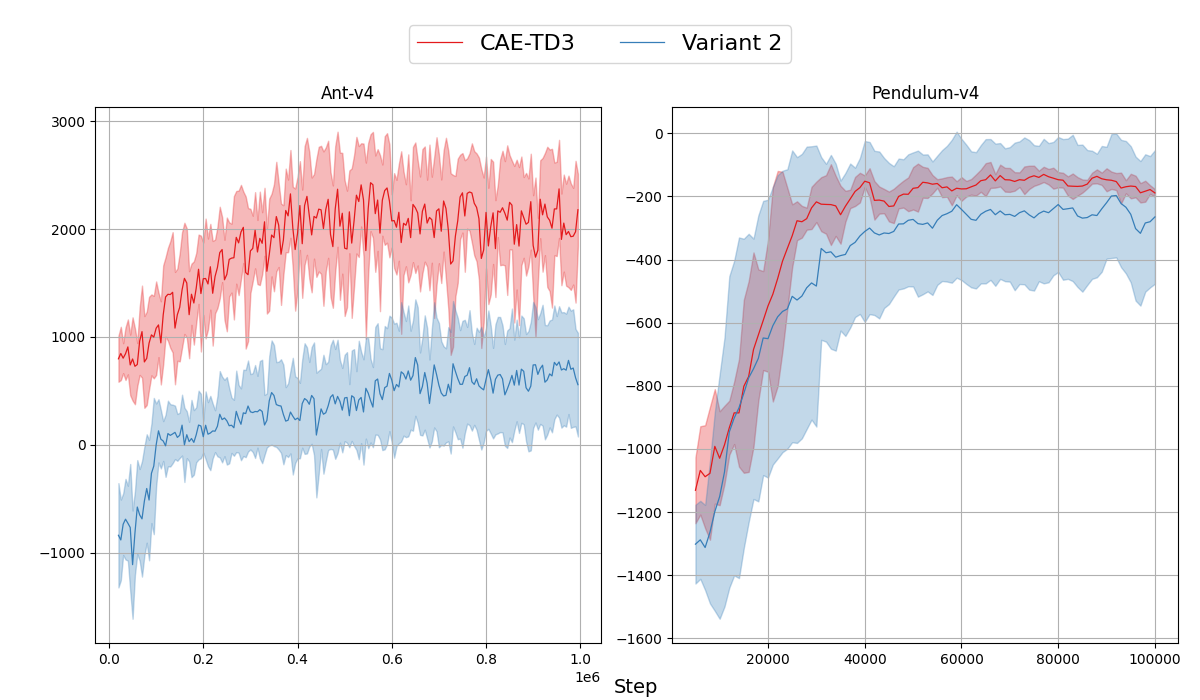}
\caption{Comparison of training performance between CAE-TD3 and Variant~2 in Ant-v4 and Pendulum-v4.}
\label{fig: cmp inc ant pendulum method 2}
\end{figure}

\texttt{Variant2}: The actor receives only historical observations, while the critic processes historical observation–action pairs. We also adopt a BERT-style schema \cite{Devlin2019BERTPO}, where the actor outputs a sequence of actions over 
$N+1$ steps, but only the final action is taken as $\bfa_t$ for interaction with the environment. Since our objective is to infer the full state from historical observations, encoding historical actions becomes potentially redundant, as they are external inputs rather than products of the environment dynamics. Consequently, action encoding is eliminated entirely, as illustrated in Figure~\ref{fig: method2}. During training, historical actions are no longer sampled from the replay buffer but instead generated by the actor, such that the critic evaluates tuples of the form $(\bar{\bfs}_t, \bar{\bfa}_t)$.

Experimental results in Figure~\ref{fig: cmp inc ant pendulum method 2} indicate that this method performs well in low-dimensional environments but fails to scale to more complex tasks. This limitation may arise because the extended optimization objective exceeds the model’s capacity, although determining the precise network size required lies beyond the scope of this work.

\begin{figure}[ht]
\centering
\includegraphics[width=0.9\linewidth]{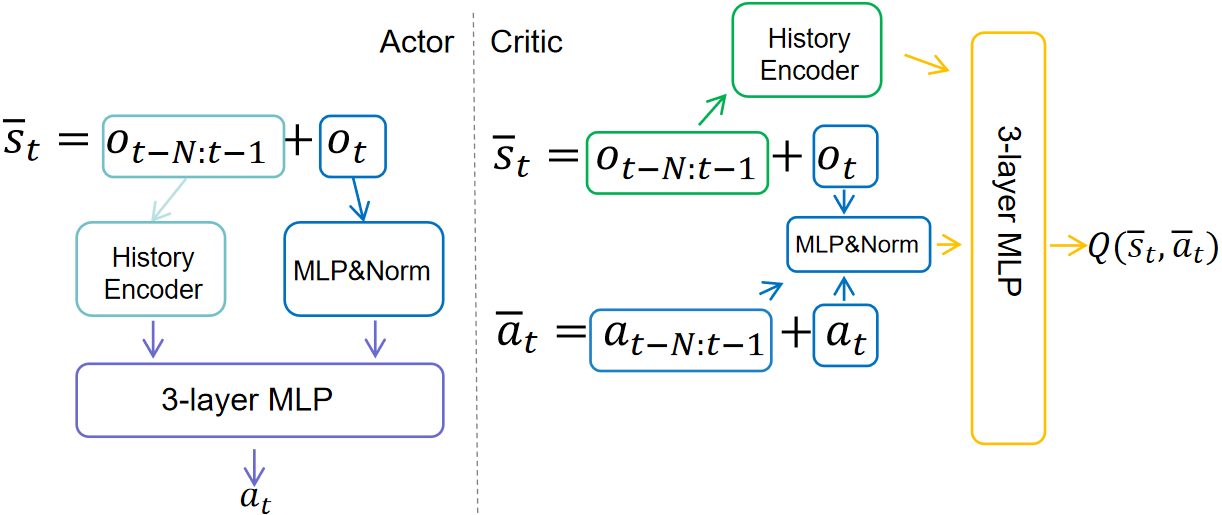}
\caption{Third variant of CAE-TD3, a simplified version of Variant~2, where the actor directly outputs $a_t$.}
\label{fig: method3}
\centering
\includegraphics[width=0.9\linewidth]{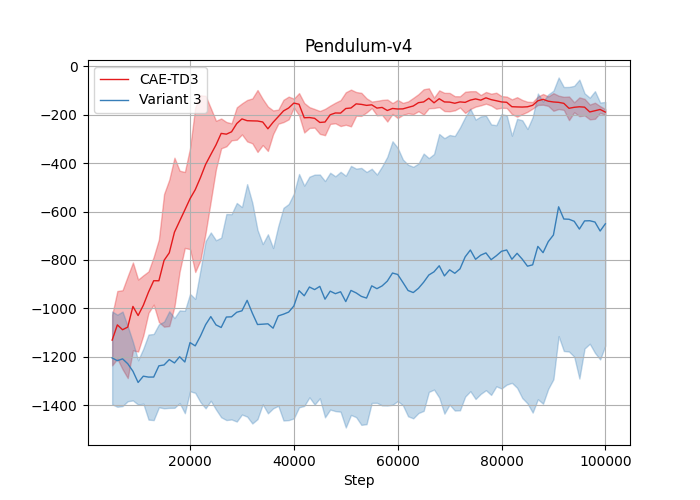}
\caption{Comparison of training performance between CAE-TD3 and Variant~3 in Pendulum-v4.}
\label{fig: cmp inc ant pendulum method 3}
\end{figure}

\texttt{Variant3}: As illustrated in Figure~\ref{fig: method3}, this design can be regarded as a simplified version of \texttt{Variant2}, where the actor’s output is reduced to $a_t$. The experimental results in Figure~\ref{fig: cmp inc ant pendulum method 3} show that this setting leads to a substantial degradation in performance. These findings suggest that encoding historical observations and actions within the same latent space is essential for achieving optimal performance.

\bibliographystyle{ieeetr}
\bibliography{a}
\end{document}